\documentclass[11pt,a4paper]{article}
\usepackage[nohyperref]{naaclhlt2019}
\aclfinalcopy
\usepackage{times}
\usepackage{latexsym}
\usepackage{url}
\usepackage{multirow}
\usepackage{rotating}
\usepackage{amsmath}
\usepackage{graphicx}
\usepackage{color}
\usepackage{caption}
\usepackage{placeins}
\usepackage{wrapfig}
\usepackage{sidecap}
\newcommand{\ignore}[1]{}
\usepackage{xargs}                      

\usepackage[colorinlistoftodos,prependcaption,textsize=tiny]{todonotes}
\newcommandx{\cjd}[2][1=]{\todo[linecolor=cyan,backgroundcolor=cyan!25,bordercolor=cyan,#1]{#2}}

\title{An Empirical Investigation of Global and Local Normalization for Recurrent Neural Sequence Models Using a Continuous Relaxation to Beam Search}
\author{Kartik Goyal \\
  School of Computer Science \\
  Carnegie Mellon University \\
  {\tt kartikgo@cs.cmu.edu} \\\And
  Chris Dyer \\
  Deepmind \\
  {\tt cdyer@google.com} \\ \And
  Taylor Berg-Kirkpatrick \\
  University of California, San Diego \\
  {\tt tberg@ucsd.edu}}

\date{}

\begin{document}
\maketitle
\begin{abstract}
  Globally normalized neural sequence models are considered superior to their locally normalized equivalents because they may ameliorate the effects of label bias. However, when considering high-capacity neural parametrizations that condition on the whole input sequence, both model classes are theoretically equivalent in terms of the distributions they are capable of representing. Thus, the practical advantage of global normalization in the context of modern neural methods remains unclear. In this paper, we attempt to shed light on this problem through an empirical study. We extend an approach for search-aware training via a continuous relaxation of beam search \cite{goyal2017continuous} in order to enable training of \emph{globally normalized} recurrent sequence models through simple backpropagation. We then use this technique to conduct an empirical study of the interaction between global normalization, high-capacity encoders, and search-aware optimization. We observe that in the context of  inexact search, globally normalized neural models are still more effective than their locally normalized counterparts. Further, since our training approach is sensitive to warm-starting with pre-trained models, we also propose a novel initialization strategy based on self-normalization for pre-training globally normalized models. We perform analysis of our approach on two tasks: CCG supertagging and Machine Translation, and demonstrate the importance of global normalization under different conditions while using search-aware training.
\end{abstract}
\section{Introduction}

Neural encoder-decoder models have been tremendously successful at a variety of NLP tasks, such as machine translation \cite{sutskever2014sequence,bahdanau2014neural}, parsing \cite{dyer2016recurrent,dyer2015transition}, summarization \cite{rush2015neural}, dialog generation \cite{serban2015building}, and image captioning \cite{xu2015show}. 
With these models, the target sequence is generated in a left-to-right step-wise manner with the predictions at every step being conditioned on the input sequence and the whole prediction history. This long-distance memory precludes exact search for the maximally scoring sequence according to the model and therefore, approximate algorithms like greedy search or beam search are necessary in practice during decoding. In this scenario, it is natural to resort to search-aware learning techniques for these models which makes the optimization objective sensitive to any potential errors that could occur due to inexact search in these models. 

This work focuses on comparison between search-aware locally normalized sequence models that involve projecting the scores of items in the vocabulary onto a probability simplex at each step and globally normalized/unnormalized sequence models that involve scoring sequences without explicit normalization at each step. When conditioned on the the full input sequence and the entire prediction history, both locally normalized and globally normalized conditional models should have same expressive power under a high-capacity neural parametrization 
in theory, as they can both model same set of distributions over all finite length output sequences \cite{smith2007weighted}. 
However, locally normalized models are constrained in how they respond to search errors during training since the scores at each decoding step must sum to one. To let a search-aware training setup have the most flexibility, abandoning this constraint may be useful for easier optimization. 

In this paper, we demonstrate that the interaction between approximate inference and non-convex parameter optimization results in more robust training and better performance for models with global normalization compared to those with the more common locally normalized parametrization. 
We posit that this difference is due to \emph{label bias} \cite{bottou-91a} arising from the interaction of approximate search and search-aware optimization in locally normalized models.  A commonly understood source of label bias in \emph{locally normalized} sequence models is an effect of conditioning only on partial input (for example, only the history of the input) at each step during decoding \cite{andor2016globally,lafferty2001conditional, wiseman2016sequence}. We discus another potential source of label bias arising from approximate search with locally normalized models that may be present even with access to the full input at each step.
To this end, we train search-aware globally and locally normalized models in an end-to-end (sub)-differentiable manner using a continuous relaxation to the discontinuous beam search procedure introduced by \citet{goyal2017continuous}. This approach requires initialization with a suitable globally normalized model to work in practice. Hence, we also propose an initialization strategy based upon self-normalization for pre-training globally normalized models.

We demonstrate the effect of both sources of label bias through our experiments on two common sequence tasks: CCG supertagging and machine translation. 
We find that label bias can be eliminated by both, using a powerful encoder, and using a globally normalized model. We observe that global normalization yields performance gains over local normalization and is able to ameliorate label bias especially in scenarios that involve a very large hypothesis space.

\section{Recurrent Sequence Models and Effects of Normalization}

We now introduce the notation that we will use in the remainder of the paper for describing locally and globally normalized neural sequence-to-sequence models.
We are interested in the probability of output sequence, $\mathbf{y}$, conditioned on input sequence, $\mathbf{x}$. Let $s(\mathbf{x},y_{1:i-1},y_i)$ be a non negative score of output label $y$ at time-step $i$ for the input $x$ and the prediction history $y_{1:i-1}$, let $V$ be the label space, and let $\mathcal{Y}_{\mathbf{x}}$ be the space of all finite sequences for $\mathbf{x}$.\footnote{For notational convenience we suppress the dependence of the score $s$ on model parameters $\theta$.} 
A neural encoder (e.g. a bidirectional LSTM) encodes information about $\mathbf{x}$ and a recurrent neural decoder generates the output $\mathbf{y}$ (typically step-by-step from left-to-right) conditioned on the encoder.

\subsection{Locally normalized models}

Under a locally normalized model $\mathcal{M}_{L}$, the probability of $\mathbf{y}$ given $\mathbf{x}$ is:
\begin{align*}
\begin{split}
p_{\mathcal{M}_{L}}(\mathbf{y}\mid \mathbf{x})= \prod_{i=1}^{n} p(y_i \mid \mathbf{x}, y_{1:i-1})=\\\prod_{i=1}^{n}\frac{s(\mathbf{x},y_{1:i-1},y_i)}{Z_{L,i}(\mathbf{x}, y_{1:i-1})}
\end{split}
\end{align*}
where $Z_{L,i}(\mathbf{x}, y_{1:i-1})= \sum_{y \in V}s(\mathbf{x},y_{1:i-1},y)$, is the local normalizer at each time step and $n$ is the number of prediction steps. Since, the local normalizer is easy to compute, likelihood maximization based training is a standard approach for training these models.
\subsection{Globally normalized models}
In contrast, under a globally normalized model $\mathcal{M}_{G}$, the probability of $\mathbf{y}$ given $\mathbf{x}$ is:
\begin{align*}
\begin{split}
p_{\mathcal{M}_{G}}(\mathbf{y}\mid \mathbf{x})= \frac{\prod_{i=1}^{n}s(\mathbf{x},y_{1:i-1},y_i)}{Z_{G}(\mathbf{x})}
\end{split}
\end{align*}
where $Z_{G}(\mathbf{x})= \sum_{\mathbf{y} \in \mathcal{Y}} \prod_{i=1}^{n}s(\mathbf{x},y_{1:i-1},y_i)$, is the global log-normalizer. $Z_{G}(\mathbf{x})$ is intractable to estimate for most problems of interest due to the large search space therefore, an exact likelihood maximization training approach is intractable for these models. 

\subsection{Label Bias with partial input}

It was shown in \citet{andor2016globally,lafferty2001conditional}, locally normalized conditional models \emph{with access to only partial input}, $x_{1:i-1}$, at each decoding step are biased towards labeling decisions with low-entropy transition probabilities at each decoding step and, as a result, suffer from a weakened ability to revise previous decisions based upon future input observations. This phenomenon has been referred to as \emph{label bias}, and presents itself as an arbitrary allocation of probability mass to unlikely or undesirable label sequences despite the presence of well-formed sequences in training data. \citet{andor2016globally} prove that this class of locally normalized models that relies on the structural assumption of access to only left-to-right partial input at each step,
\begin{align*}
    \prod_{i=1}^{n} p(y_i \mid \mathbf{x}, y_{1:i-1}) = \prod_{i=1}^{n} p(y_i \mid x_{1:i-1}, y_{1:i-1}),
\end{align*}
is strictly less expressive than its globally normalized counterpart. 

However, the standard sequence-to-sequence models used most often in practice and presented in this paper actually condition the decoder on a summary representation of the \emph{entire input sequence}, $\mathbf{x}$, computed by a neural encoder. Hence, depending on the power of the encoder, it is commonly thought that such models avoid this type of label bias. For these models, both locally normalized and globally normalized conditional models are equally expressive, in principle, with a sufficiently powerful encoder. 

However, as we suggest in the next section and show empirically in experiments, this does not necessarily mean that both parametrizations are equally amenable to gradient-based training in practice, particularly when the search space is large and search-aware training techniques are used. We will argue that they suffer from a related, but distinct, form of bias introduced by inexact decoding.

\subsection{Search-aware training}
To improve performance with inexact decoding methods (e.g. beam search), search-aware training techniques take into account the decoding procedure that will be used at test time and adjust the parameters of the model to maximize prediction accuracy under the decoder. Because of the popularity of beam search as a decoding procedure for sequence models, in this paper we focus on beam search-aware training. While many options are available, including beam-search optimization (BSO) \cite{wiseman2016sequence}, in Section~\ref{soft-beams} we will describe the particular search-aware training strategy we use in experiments \cite{goyal2017continuous}, chosen for its simplicity.


\subsection{Label Bias due to approximate search}

We illustrate via example how optimization of locally normalized models may suffer from a new kind of label bias when using beam search-aware training, and point to reasons why this issue might be mitigated by the use of globally normalized models. While the scores of successors of a single candidate under a locally normalized model are constrained to sum to one, scores of successors under a globally normalized model need only be positive. Intuitively, during training, this gives the globally normalized model more freedom to downweight undesirable intermediate candidates in order avoid search errors.  
\begin{figure}[t]
\centering
\includegraphics[width=0.5\textwidth]{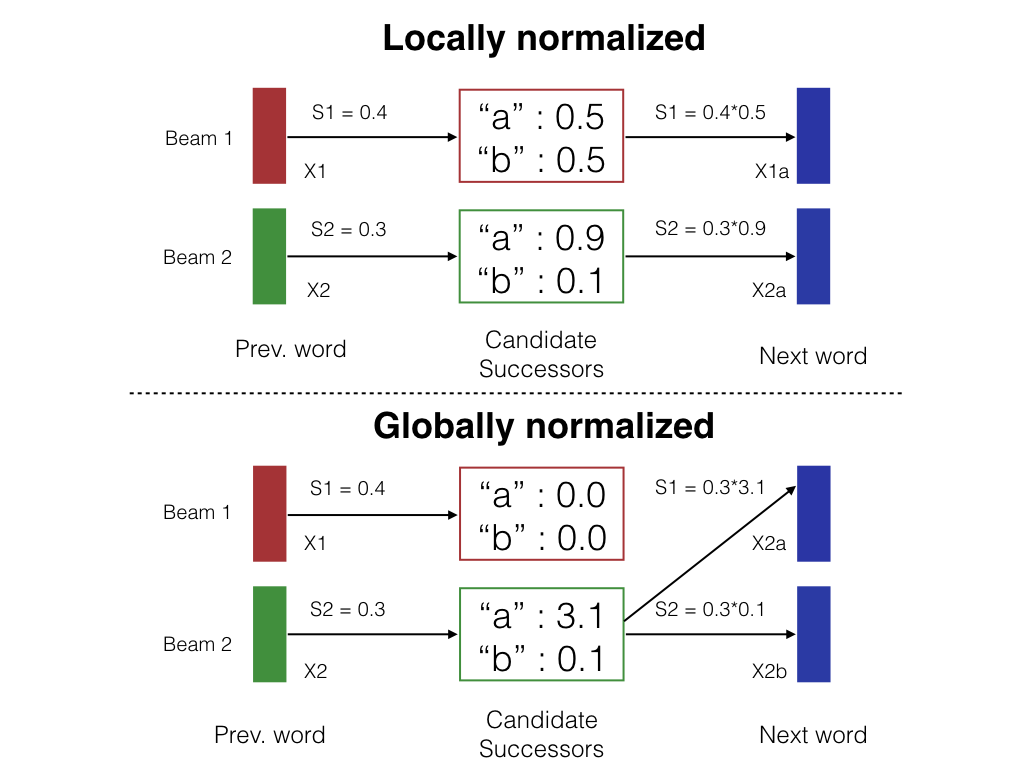}
\vspace{-0.2cm}
\caption{\label{lbias} Illustrative example of bias arising in locally normalized models due to beam search. Red indicates the candidate that optimization should learn to discard and green indicates the candidate that should be propagated. Locally normalized models are constrained to return normalized scores for the successors of each candidate, while globally normalized models are unconstrained and can more easily learn to drop successors of the red candidate. 
}
\vspace{-0.2cm}
\end{figure}
\begin{figure*}[t]
\centering
\includegraphics[width=0.9\textwidth]{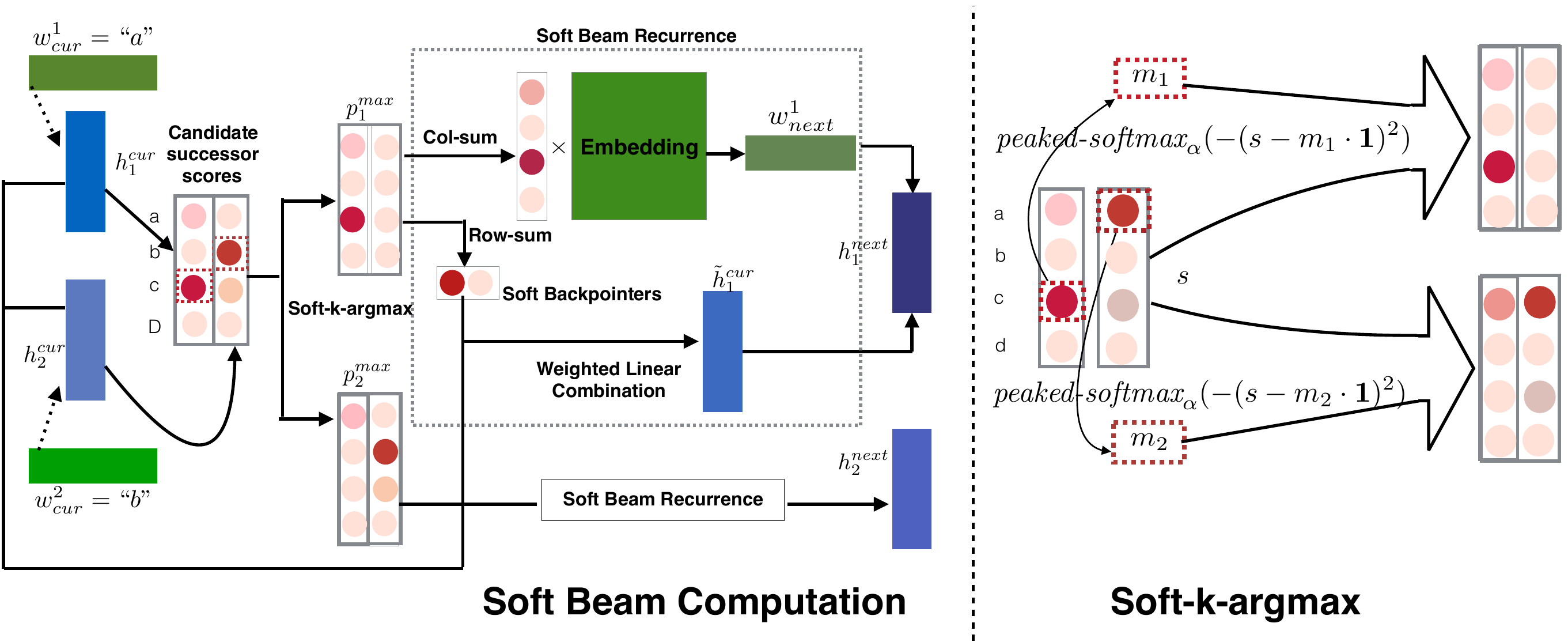}
\vspace{-0.2cm}
  \caption{\label{softbeam} {\bf Left:} Computing LSTM hidden states at a subsequent step using continuous relaxation to beam search for beam size of 2. {\bf Right:} Illustration of top-k-argmax. $m_1$ and $m_2$ are the maximum and second maximum candidate scores. }
  \vspace{-0.2cm}
\end{figure*}

In the example beam search decoding problem in Figure~\ref{lbias}, we compare the behavior of locally and globally normalized models at a single time step for a beam size of two. 
In this example, we assume that the score for beams in  both the models is  exactly the same until the step shown in Figure~\ref{lbias}.
Suppose that the lower item on the beam(X2) is correct, and thus, for more effective search, we would prefer the models scores to be such that only successors of the lower beam item are present on the beam at the next step. However, since, the scores at each step for a locally normalized model are constrained to sum to one, the upper beam item(X1) generates successors with scores comparable to those of the lower beam item. As we see in the example, due to the normalization constraint, search-aware training of the locally normalized model might find it difficult to set the parameters to prevent extension of the poorer candidate. In contrast, because the scores of a \emph{globally normalized} model are not constrained to sum to one, the parameters of the neural model can be set such that \emph{all} the successors of the bad candidate have a very low score and thus do not compete for space on the beam. This illustrates a mechanism by which search-aware training of globally normalized models in a large search spaces might be more effective. However as discussed earlier, if we can perform \emph{exact search} then this \emph{label bias} ceases to exist because both the models have the same expressive power with a search-agnostic optimization scheme. In experiments, we will explore this trade-off empirically.

\section{Search-aware Training for Globally Normalized Models}
In order to conduct an empirical study with meaningful comparisons, we devise an extension of the relaxed beam-search based optimization proposed by \citet{goyal2017continuous} that allows us to train both the search-aware globally and locally normalized models in a similar manner with the same underlying architecture.
\subsection{\label{soft-beams} Continuous Relaxation to Beam Search}
 Following \citet{goyal2017continuous}, we train a beam-search aware model by optimizing a continuous surrogate approximation to a \emph{direct loss} objective, $J$, defined as a function of the output of beam search and the ground truth sequence $\mathbf{y^*}$:
\begin{align*}
\min_{\theta} J(\mathbf{x},\theta,\mathbf{y^*}) = \min_{\theta} \ell(\textit{Beam}(\mathbf{x},\mathcal{M}(\theta)),\mathbf{y^*})
\end{align*}
Here $\ell$ is a function that computes the loss of the model's prediction produced by beam search 
, and $\mathcal{M}$ refers to the model parametrized by $\theta$. While this objective is search-aware, it is discontinuous and difficult to optimize because beam search involves discrete \emph{k-argmax} operations. Therefore, \citet{goyal2017continuous} propose a continuous surrogate, $\tilde{J}$, by defining a continuous approximation (\emph{soft-k-argmax}) of the discrete \emph{k-argmax} and using this to compute an approximation to a composition of the loss function and the beam search function.
\begin{align*}
\min_\theta \tilde{J}(\mathbf{x},\theta,\mathbf{y^*}) \approx \min_\theta (\ell \circ Beam)(\mathbf{x}, \mathcal{M}(\theta), \mathbf{y^*})
\end{align*}
 The \emph{soft-k-argmax} procedure involves computing distances between the scores of the successors and the $k^{th}$-max score and using the temperature based argmax operation \cite{maddison2016concrete,jang2016categorical,goyal2017differentiable} to get an output peaked on the $k^{th}$-max value as shown in the right panel of Figure~\ref{softbeam}. The temperature is a hyperparameter which is typically annealed toward producing low entropy distributions during optimization. As shown in the left panel of Figure~\ref{softbeam}, the \emph{soft candidate vectors} and the \emph{soft backpointers} are computed  at every decoding step using this \emph{soft-k-argmax} operation in order to generate the embeddings and recurrent hidden states of the LSTM at each step of the \emph{soft beam search} procedure. With a locally decomposable loss like Hamming loss, both \emph{soft loss} and \emph{soft scores} for the relaxed procedure are iteratively computed so that the end-to-end objective computation can be described by a computation graph that is amenable to backpropagation.

Using this relaxation, point-wise convergence of the surrogate objective to the original objective can be established ($\alpha$ is the inverse temperature):
\begin{align*}
\tilde{J}_\alpha(\mathbf{x}, \theta, \mathbf{y^*}) \xrightarrow[\textit{p}]{\alpha \to \infty} J(\mathbf{x},\theta,\mathbf{y^*})
\end{align*}
\citet{goyal2017continuous} demonstrated empirically that optimizing the surrogate objective, $\tilde{J}$ -- which can be accomplished via simple backpropagation for decomposable losses like Hamming distance -- leads to improved performance at test time.



In experiments, for training locally normalized models,  we use \emph{log-normalized} successor scores. However, for training globally normalized models, we will directly use \emph{unnormalized scores}, which are $\in \mathrm{R}_{+}$.

\subsection{Initialization for training globally normalized models}
\citet{goyal2017continuous} reported that \emph{initialization} with a locally normalized model pre-trained with teacher-forcing was important for their continuous beam search based approach to be stable and hence they used the locally normalized log-scores for their search-aware training model. In this work, we experimented with the unnormalized candidate successor scores and found that initializing the optimization for a globally normalized objective with a cross-entropy trained locally normalized model resulted in unstable training. This is expected because the locally normalized models are parametrized in a way such that using the scores before the softmax normalization results in a very different outcome than using scores after local normalization. For example, the locally normalized Machine Translation model in Table~\ref{logz}, that gives a BLEU score of $27.62$ when decoded with beam search using locally normalized scores, results in BLEU of $4.30$ when beam search decoding is performed with unnormalized scores. Pre-training a truly globally normalized model for initialization is not straghtforward because no exact likelihood maximization techniques exist for globally normalized models as the global normalizer is intractable to compute. 

Therefore, we propose a new approach to initialization for search-aware training of globally normalized models: we pre-train a locally normalized model that is parametrized like a globally normalized model. More specifically, we train a locally normalized model with its distribution over the output sequences denoted by $p_L(\mathcal{Y})$ such that we can easily find a globally normalized model with a distribution $p_G(\mathcal{Y})$ that matches $p_L(\mathcal{Y})$. Following the notation in Section 2, for a locally normalized model, the log-probability of a sequence is: 
\begin{align*}
    \sum_{i=1}^{n} \left[\log s(\mathbf{x},y_{1:i-1},y_i) - \log Z_{L,i}(\mathbf{x}, y_{1:i-1})\right]
\end{align*}
and for a globally normalized model it is:
\begin{align*}
     \left[\sum_{i=1}^{n} \log s(\mathbf{x},y_{1:i-1},y_i)\right] - \log Z_{G}(\mathbf{x})
\end{align*}
\subsubsection{Self Normalization}
One way to find a locally normalized model that is parametrized like a globally normalized model is to ensure that the local normalizer at each step, $\log~Z_{L,i}(\mathbf{x}, y_{1:i-1})$, is $0$. With the local normalizer being zero it is straightforward to see that the log probability of a sequence under a locally normalized model can easily be interpreted as log probability of the sequence under a globally normalized model with the global log-normalizer, $\log~Z_{G}(\mathbf{x}) = 0$. This training technique is called \emph{self-normalization} \cite{andreas2015and} because the resulting models' unnormalized score at each step lies on a probability simplex. A common technique for training self-normalized models is L2-regularization of local log normalizer which encourages learning a model with $\log~Z = 0$ and was found to be effective for learning a language model by \citet{devlin2014fast}\footnote{Noise Contrastive Estimation \cite{mnih2012fast, gutmann2010noise} is also an alternative to train unnormalized models but our experiments with NCE were unstable and resulted in worse models.}. 
The L2-regularized cross entropy objective is given by:
\begin{align*}
    \min_\theta \sum_{\mathbf{x},\mathbf{y^*}\in \mathcal{D}}-\sum_{i=1}^{n} \log~p(y_i^*\mid \mathbf{x}, y_{1:i-1})\\ + \lambda \cdot (\log~Z_{L,i}(\mathbf{x}, y_{1:i-1}))^2
\end{align*}
In Table~\ref{logz}, we report the mean and variance of the local log normalizer on the two different tasks using L2-regularization (\emph{L2}) based self normalization and no self normalization (\emph{CE}). We observe that \emph{L2} models are competitive performance-wise to the cross-entropy trained locally normalized models while resulting in a much smaller local log-normalizer on average. Although, we couldn't minimize $\log~Z$ exactly to 0, we observe in Section 4 that this is sufficient to train a reasonable initializer for the search-aware optimization of globally normalized models.
It is important to note that these approaches yield a \emph{globally normalized} model that is equivalent to a locally normalized model trained via teacher-forcing and hence these are only used to \emph{warm-start} the search-aware optimization of globally normalized models. Our search-aware training approach is free to adjust the parameters of the models such that the final globally normalized model has a non-zero log-normalizer $Z_{G}$ over the data.
\begin{table}[h]
\centering
\small
\begin{tabular}{|l|l|l|l|l|l|l|}
\hline
\multicolumn{2}{|l|}{}    & \multicolumn{2}{l|}{Train logZ} &      \multicolumn{2}{l|}{Dev logZ} & \multirow{2}{*}{\begin{tabular}[c]{@{}l@{}}Acc/\\ BLEU\end{tabular}} \\ \cline{1-6}
\multicolumn{2}{|l|}{}    & Mean                                                  & Var   & Mean          & Var           &                                                                      \\ \hline
\multicolumn{1}{|c|}{\multirow{2}{*}{CCG}} & CE & 21.08                                                 & 9.57  & 21.96         & 9.18          & 93.3                                                                 \\ \cline{2-7} 
\multicolumn{1}{|c|}{}                     & L2 & 0.6                                                   & 0.29  & 0.26          & 0.08          & 91.9                                                                 \\ \hline
\multirow{2}{*}{MT}                        & CE & 24.7                                                  & 115.4 & 25.8          & 129.1         & 27.62                                                                \\ \cline{2-7} 
                                           & L2 & 0.65                                                  & 0.18  & 0.7           & 0.29          & 26.63                                                                \\ \hline
\end{tabular}
\caption{\label{logz} Comparison of logZ between cross entropy trained models (CE) and self normalized models (L2) for CCG supertagging and Machine Translation tasks.}
\end{table}

Other possible approaches to project locally normalized models onto globally normalized models include distribution matching via knowledge distillation \cite{hinton2015distilling}.
We leave exploration of warm-starting of search aware optimization with this approach to future work.
\section{Experiments and Empirical Analysis}
To empirically analyze the interaction between label bias arising from different sources, search-aware training, and global normalization, we conducted experiments on two tasks with vastly different sizes of output space: CCG supertagging and Machine Translation. As described in the next section, the task of tagging allows us to perform controlled experiments which explicitly study the effect of amount of input information available to the decoder at each step, we analyze the scenarios in which search aware training and global normalization are expected to improve the model performance.

In all our experiments, we report results on training with standard teacher forcing optimization and self-normalization as our baselines. We report results with both search-aware locally and globally normalized models (Section~3.1) after warm starting with both cross entropy trained models and self-normalized models to study the effects of search-aware optimization and global normalization. We follow \citet{goyal2017continuous} and use the decomposable Hamming loss approximation with search-aware optimization for both the tasks and decode via \emph{soft beam search decoding} method which involves continuous beam search with soft backpointers for the LSTM Beam search dynamics as described in Section 3, but using identifiable backpointers and labels (using MAP estimates of soft backpointers and labels) to decode.

We tune hyperparameters like learning rate and annealing schedule by observing performance on development sets for both the tasks. We performed at least three random restarts for each class and report results based on best development performance. 
\subsection{CCG supertagging}
We used the standard splits of CCG bank \cite{hockenmaier2002acquiring} for training, development, and testing. The label space of supertags is 1,284 and the labels are correlated with each other based on their syntactic relations. The distribution of supertag labels in the training data exhibits a long tail distribution. This task is sensitive to the long range sequential decisions because it encodes rich syntactic information about the sentence. Hence, this task is ideal to analyze the effects of label bias and search effects. We perform minor preprocessing on the data similar to the preprocessing in \citet{vaswani2016supertagging}. For experiments related to search aware optimization, we report results with beam size of 5.\footnote{We observed similar results with beam size 10}
\subsubsection{Tagging model for ablation study}
We changed the standard sequence-to-sequence model to be more suitable for the tagging task. This change also lets us perform controlled experiments pertaining to the amount of input sequence information available to the decoder at each time step. 

In a standard encoder-decoder model with attention, the initial hidden state of the decoder is often some function of the final encoder state so that the decoder's predictions can be conditioned on the full input. For our tagging experiments, instead of influencing the initial decoder state with the encoder, we set it to a vector of zeros. Thus the information about input for prediction is \emph{only} available via the attention mechanism. In addition to the change above, we also forced the model to attend to only the $i^{th}$ input representation while predicting the $i^{th}$ label. This is enforceable because the output length is equal to the input length and it is also a more suitable structure for a tagging model. With these changes in the decoder, we can precisely control the amount of information about the input available to the decoder at each prediction step. For example, with a unidirectional LSTM encoder, the decoder at $i^{th}$ step only has access to input till the $i^{th}$ token and the prediction history:
\begin{align*}
    p(y_i \mid \mathbf{x}, y_{1:i-1}) = p(y_i \mid x_{1:i}, y_{1:i-1})
\end{align*}
This setting lets us clearly explore the classical notion of \emph{label bias} arising out of access to partial input at each prediction step (Section 2.3). A bidirectional LSTM encoder, however provides access to all of the input information to the decoder at all the prediction steps.
\begin{table}[h]
\small
\centering
\begin{tabular}{|l|l|l|}
\hline
\multicolumn{1}{|l|}{}       & Unidirectional   & Bidirectional   \\ \hline
                     pretrain-greedy    & 76.54 & 92.59       \\ \hline
                     pretrain-beam    & 77.76 & 93.29       \\ \hline
                     locally normalized  & 83.9  & \bf{93.76}     \\ \hline 
                    globally normalized & \bf{83.93} & 93.73    \\ \hline
\end{tabular}
\caption{\label{ccgce} {\bf Accuracy results on CCG supertagging when initialized with a regular teacher-forcing model}. Reported using \emph{Unidirectional} and \emph{Bidirectional} encoders respectively with fixed attention tagging decoder. \emph{pretrain-greedy} and \emph{pretrain-beam} refer to the output of decoding the initializer model. \emph{locally normalized} and \emph{globally normalized} refer to search-aware soft-beam models }
\end{table}
\begin{table}[h]
\small
\centering
\begin{tabular}{|l|l|l|}
\hline
\multicolumn{1}{|l|}{}       & Unidirectional   & Bidirectional   \\ \hline
                     pretrain-greedy    & 73.12 & 91.23       \\ \hline
                     pretrain-beam    & 73.83 & 91.94       \\ \hline
                     locally normalized  & 83.35  & \bf{92.78}     \\ \hline 
                    globally normalized & \bf{85.50} & 92.63    \\ \hline
\end{tabular}
\caption{\label{ccgl2} {\bf Accuracy results on CCG supertagging when initialized with a self normalized model}. }
\end{table}
\subsection{Machine Translation}
 We use the same dataset (the German-English portion of the IWSLT 2014 machine translation evaluation campaign \cite{cettolo2014report}), preprocessing and data splits as \citet{ranzato2015sequence} for our Machine Translation experiments. The output label/vocabulary size is 32000 and unlike tagging, the length of output sequences cannot be deterministically determined from the length of the input sequence. Moreover, the output sequence does not necessarily align monotonically with the input sequence. Hence the output sequence space for MT is much larger than that for tagging and the effects of inexact search on optimization are expected to be even more apparent for MT. We use a standard LSTM-based encoder/decoder model with a standard attention mechanism \cite{bahdanau2016task} for our MT experiments. For search-aware optimization experiments, we report results with beam size 3.\footnote{We observed similar results beam size of 5.}
\begin{table}[]
\small
\centering
\begin{tabular}{|l|l|l|}
\hline
Init-scheme $\rightarrow$ & Regular & Self-normalized \\ \hline
pretrain-greedy  & 26.24 & 25.42\\ \hline
pretrain-beam & 27.62 & 26.63\\ \hline
locally-normalized & \bf{29.28} & 27.71\\ \hline 
globally-normalized & 26.24  & \bf{29.27}\\ \hline
\end{tabular}
\caption{\label{mt} {\bf BLEU results on de-en Machine Translation.} \emph{Regular} and \emph{Self-normalized} refer to the initization scheme for soft beam search training. \emph{pretrain-greedy} and \emph{pretrain-beam} refer to the output of decoding the initializer model. \emph{locally normalized} and \emph{globally normalized} refer to search-aware soft-beam models}
\end{table}
\subsection{Results and Analysis}
The results reported in Tables~\ref{ccgce}, \ref{ccgl2} and \ref{mt} allow us to analyze the effect of interaction of label bias, inexact search and global normalization in detail.\\[1.2pt]
\subsubsection{Label bias with partial input}First, we analyze the effect of label bias that arises from conditioning on partial input (Section 2.3) during decoding on optimization of the models. The unidirectional encoder based tagging experiments suggest that conditioning on partial input during decoding results in poor models when trained with cross entropy based methods. Interestingly, all techniques improve upon this: (i) search-aware locally and globally normalized models are able to train for accuracy directly and eliminate exposure bias that arises out of the mismatch between train-time and test-time prediction methods, and, (ii) the bidirectional tagging model which provides access to all of the input is powerful enough to learn a complex relationship between the decoder and the input representations for the search space of the CCG supertagging task and results in a much better performance.\\[1.2pt]
\subsubsection{Initialization of search-aware training} Next, we analyze the importance of appropriate initialization of search-aware optimization with pretrained models. Across all the results in Tables~\ref{ccgce}, \ref{ccgl2} and \ref{mt}, we observe that search-aware optimization for locally normalized models always improves upon the pre-trained locally normalized models used for initialization. But when the search-aware optimization for globally normalized models is initialized with locally normalized CE models, the improvement is not as pronounced and in the case of MT, the performance is actually \emph{hurt} by the improper initialization for training globally normalized models -- probably a consequence of large search space associated with MT and incompatibility between unnormalized scores for search-aware optimization and locally normalized scores of the \emph{CE} model used for pre-training. When the \emph{self-normalized} models are used for initialization, optimization for globally normalized models always improves upon the pre-trained self-normalized model. It is interesting to note that we see improvements for the globally normalized models even when $log Z$ is not exactly reduced to $0$ indicating that the scores used for search-aware training initially are comparable to the scores of the pre-trained self-normalized model. We also observe that self-normalized models perform slightly worse than CE-trained models but search aware training for globally normalized models improves the performance significantly.\\[1.2pt]
\subsubsection{Search-aware training} Next, we analyze the effect of search-aware optimization on the performance of the models. Search-aware training with locally normalized models improves the performance significantly in \emph{all} our experiments which indicates that accounting for exposure bias and optimizing for predictive performance directly is important.
We also observe that the bidirectional model for tagging is quite powerful and seems to account for both \emph{exposure bias} and \emph{label bias} to a large extent. We reckon that this may be because the greedy decoding itself is very close to \emph{exact search} for this well-trained tagging model over a search space that is much simpler than that associated with MT. Therefore, the impact of search-aware optimization on the bidirectional tagger is marginal. However, it is much more pronounced on the task of MT.\\[1.2pt]
\subsubsection{Global normalization and label bias} We analyze the importance of training globally normalized models. In the specific setup for tagging with the unidirectional encoder, globally normalized models are actually \emph{more expressive} than the locally normalized models \cite{andor2016globally} as described in Section 2.3 and this is reflected in our experiments (table~\ref{ccgl2}) with tagging. The globally normalized model (warm started with a self-normalized model) performs the best among all the models in the unidirectional tagger case which indicates that it is ameliorating something beyond exposure bias which is fixed by the search-aware locally normalized model.\\
For MT (table~\ref{mt}), both globally normalized and locally normalized models are equally expressive in theory because the decoder is conditioned on the full input information at each step, but we still observe that the globally normalized model improves significantly over the self-normalized pre-trained model and the search-aware locally normalized model. This indicates that it might be ameliorating the label bias associated with inexact search (discussed in Section 2.5). As discussed in Section 3.2, the globally normalized model, when initialized with a CE trained model, performs worse because of improper initialization of the search aware training. The self-normalized model starts off 1 BLEU point worse than the CE model point but global normalization, initialized with the self-normalized model improves the performance and is competitive with the best model for MT. This suggests that a better technique for initializing the optimization for globally normalized models should be helpful in improving the performance.\\[1.2pt]
\subsubsection{Global normalization and sentence length}
In tables~\ref{bleu} and \ref{len}, we analyze the source of improvement from global normalization for MT. In table~\ref{bleu}, we report the ngram overlap scores and ratio of length of the predictions to length of hypothesis for the case when the search-aware training is initialized with a self-normalized model. We observe that the globally normalized model produces longer predictions than the locally normalized model. More interestingly, it seems to have better 3 and 4-gram overlap and slightly worse unigram and bigram overlap score than the locally normalized model. These observations suggest that globally normalized models are better able to take longer range effects into account and are also cautious about predicting the \emph{end-of-sentence} symbol too soon. Moreover, in table~\ref{len}, we observe that globally normalized models perform better on all the length ranges but especially so on long sentences.
\begin{table}[]
\small
\centering
\begin{tabular}{|l|l|l|}
\hline
 & N-gram overlap & Length ratio \\ \hline
pretrain-beam & 63.5/35.7/21.8/13.7 & 0.931\\ \hline
locally-normalized & 66.9/39.4/22.7/14.0 & 0.918\\ \hline 
globally-normalized & 65.0/39.1/23.2/14.7  & 0.959\\ \hline
\end{tabular}
\caption{\label{bleu} {\bf Breakdown of BLEU results on de-en Machine Translation dev set.} Reported on Self-normalized initialization}
\end{table}
\begin{table}[]
\small
\centering
\begin{tabular}{|l|l|l|l|l|}
\hline
Src sent-length $\rightarrow$ & \bf{0-20} & \bf{20-30} & \bf{30-40} & \bf{40+}  \\ \hline
pretrain-beam & 29.36 & 25.73 & 24.71 & 24.50\\ \hline
locally-normalized & 32.35 & 26.95 & 25.39 & 25.2\\ \hline 
globally-normalized & 33.21  & 28.08 & 26.75 & 26.41\\ \hline
\end{tabular}
\caption{\label{len} {\bf BLEU scores with different length inputs on dev set} Reported on Self-normalized initialization. The header specifies the range of length of the input sentences}
\end{table}
\section{Related Work}
Much of the existing work on search-aware training of globally normalized neural sequence models uses some mechanism like early updates \cite{collins2004incremental} that relies on explicitly tracking if the gold sequence falls off the beam and is not end-to-end continuous. \citet{andor2016globally} describe a method for training globally normalized neural feedforward models, which involves optimizing a CRF-based likelihood where the normalizer is approximated by the sum of the scores of the final beam elements. They describe label bias arising out of conditioning on partial input and hence focused on the scenario in which locally normalized models can be less expressive than globally normalized models, whereas we also consider another source of label bias which might be affecting the optimization of equally expressive locally and globally normalized conditional models. 
\citet{wiseman2016sequence} also propose a beam search based training procedure that uses unnormalized scores similar to our approach. Their models achieve good performance over CE baselines -- a pattern that we observe in our results as well. In this work, we attempt to empirically analyze the factors affecting this boost in performance with end-to-end continuous search-aware training \cite{goyal2017continuous} for globally normalized models.

\citet{smith2007weighted} proved that locally normalized conditional PCFGs and unnormalized conditional WCFGs are equally expressive for finite length sequences and posit that Maximum Entropy Markov Models (MEMMs) are weaker than CRFs because of the structural assumptions involved with MEMMs that result in label bias. 

Recently, energy based neural structured prediction models \cite{amos2016input, belanger2016structured, belanger2017end} were proposed that define an energy function over candidate structured output space and use gradient based optimization to form predictions making the overall optimization search aware. These models are designed to model global interactions between the output random variables without specifying strong structural assumptions. 
\section{Conclusion}
We performed empirical analysis to analyze the interaction between label bias, search-aware optimization and global normalization in various scenarios. We proposed an extension to the continuous relaxation to beam search proposed by \citet{goyal2017continuous} to train search-aware globally normalized models and comparable locally normalized models. We find that in the context of inexact search over large output spaces, globally normalized models are more effective than the locally normalized models in spite of them being equivalent in terms of their expressive power. 
\section*{Acknowledgement}
This project is funded in part by the NSF under grant 1618044. We thank the three anonymous reviewers for their helpful feedback.
\bibliographystyle{acl_natbib_nourl}
\bibliography{refs}

\end{document}